\documentclass[10pt,twocolumn,letterpaper]{article}

\usepackage{cvpr}
\usepackage{times}
\usepackage{epsfig}
\usepackage{graphicx}
\usepackage{amsmath}
\usepackage{amssymb}
\usepackage{subcaption}
\usepackage{cuted}
\usepackage{capt-of}

\usepackage[breaklinks=true,bookmarks=false]{hyperref}
\hypersetup{
    colorlinks=true,
    linkcolor=red,
    filecolor=magenta,      
    urlcolor=magenta,
}

\cvprfinalcopy 

\def\cvprPaperID{1949*} 

\ifcvprfinal\fi
\begin{document}

\title{UnrealText: Synthesizing Realistic Scene Text Images from the Unreal World}

\author{Shangbang Long\\
Carnegie Mellon University\\
{\tt\small shangbal@cs.cmu.edu}
\and
Cong Yao\\
Megvii (Face++) Technology Inc. \\
{\tt\small yaocong2010@gmail.com}
}

\maketitle

\begin{strip}\centering
\includegraphics[width=1.0\textwidth]{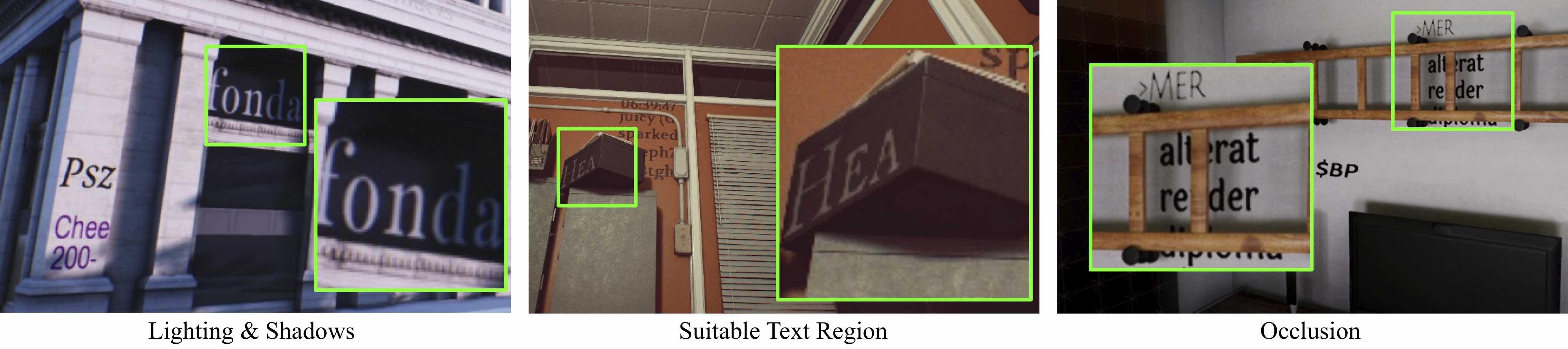}
\vspace{-7mm}
\captionof{figure}{Demonstration of the proposed \textbf{UnrealText} synthesis engine, which achieves photo-realistic lighting conditions, finds suitable text regions, and realizes natural occlusion (from \textit{left} to \textit{right}, zoomed-in views marked with \textcolor{green}{green squares}).
\label{fig:demo}}
\vspace{-3mm}
\end{strip}

\begin{abstract}
 Synthetic data has been a critical tool for training scene text detection and recognition models. 
 On the one hand, synthetic word images have proven to be a successful substitute for real images in training scene text recognizers. 
 On the other hand, however, scene text detectors still heavily rely on a large amount of manually annotated real-world images, which are expensive. 
 In this paper, we introduce UnrealText, an efficient image synthesis method that renders realistic images via a 3D graphics engine. 
 3D synthetic engine provides realistic appearance by rendering scene and text as a whole, and allows for better text region proposals with access to precise scene information, e.g. normal and even object meshes. 
 The comprehensive experiments verify its effectiveness on both scene text detection and recognition. 
 We also generate a multilingual version for future research into multilingual scene text detection and recognition. 
 Additionally, we re-annotate scene text recognition datasets in a case-sensitive way and include punctuation marks for more comprehensive evaluations. 
 The code and the generated datasets are released at: 
 \href{https://jyouhou.github.io/UnrealText/}{https://jyouhou.github.io/UnrealText/}. 
\end{abstract}

\vspace{-4mm}

\section{Introduction}

With the resurgence of neural networks, the past few years have witnessed significant progress in the field of scene text detection and recognition. 
However, these models are data-thirsty, and it is expensive and sometimes difficult, if not impossible, to collect enough data. 
Moreover, the various applications, from traffic sign reading in autonomous vehicles to instant translation, require a large amount of data specifically for each domain, further escalating this issue. 
Therefore, synthetic data and synthesis algorithms are important for scene text tasks. 
Furthermore, synthetic data can provide detailed annotations, such as character-level or even pixel-level ground truths that are rare for real images due to high cost. 

Currently, there exist several synthesis algorithms~\cite{wang2012end,jaderberg2014synthetic,gupta2016synthetic,zhan2018verisimilar} that have proven beneficial. 
Especially, in scene text recognition, training on synthetic data~\cite{jaderberg2014synthetic,gupta2016synthetic} alone has become a widely accepted standard practice. 
Some researchers that attempt training on both synthetic and real data only report marginal improvements~\cite{li2018show,long2019alchemy} on most datasets. 
Mixing synthetic and real data is only improving performance on a few difficult cases that are not yet well covered by existing synthetic datasets, such as seriously blurred or curved text. 
This is reasonable, since cropped text images have much simpler background, and synthetic data enjoys advantages in larger vocabulary size and diversity of backgrounds, fonts, and lighting conditions, as well as thousands of times more data samples. 

On the contrary, however, scene text detection is still heavily dependent on real-world data. 
Synthetic data~\cite{gupta2016synthetic,zhan2018verisimilar} plays a less significant role, and only brings marginal improvements. 
Existing synthesizers for scene text detection follow the same paradigm. 
First, they analyze background images, e.g. by performing semantic segmentation and depth estimation using off-the-shelf models. 
Then, potential locations for text embedding are extracted from the segmented regions. 
Finally, text images (foregrounds) are blended into the background images, with perceptive transformation inferred from estimated depth. 
However, the analysis of background images with off-the-shelf models may be rough and imprecise. 
The errors further propagate to text proposal modules and result in text being embedded onto unsuitable locations. 
Moreover, the text embedding process is ignorant of the overall image conditions such as illumination and occlusions of the scene. 
These two factors make text instances outstanding from backgrounds, leading to a gap between synthetic and real images.

In this paper, we propose a synthetic engine that synthesizes scene text images from 3D virtual world. 
The proposed engine is based on the famous \textit{Unreal Engine 4 (UE4)}, and is therefore named as \textit{UnrealText}. 
Specifically, text instances are regarded as planar polygon meshes with text foregrounds loaded as texture. 
These meshes are placed in suitable positions in 3D world, and rendered together with the scene as a whole. 

As shown in Fig. \ref{fig:demo}, the proposed synthesis engine, by its very nature, enjoys the following advantages over previous methods: 
(1) Text and scenes are rendered together, achieving realistic visual effects, e.g. illumination, occlusion, and perspective transformation. 
(2) The method has access to precise scene information, e.g. normal, depth, and object meshes, and therefore can generate better text region proposals. 
These aspects are crucial in training detectors. 

To further exploit the potential of UnrealText, we design three key components: (1) A view finding algorithm that explores the virtual scenes and generates camera viewpoints to obtain more diverse and natural backgrounds.
(2) An environment randomization module that changes the lighting conditions regularly, to simulate real-world variations. 
(3) A mesh-based text region generation method that finds suitable positions for text by probing the 3D meshes. 

The contributions of this paper are summarized as follows: 
(1) We propose a brand-new scene text image synthesis engine that renders images from 3D world, which is entirely different from previous approaches that embed text on 2D background images, termed as \textbf{UnrealText}. 
The proposed engine achieves realistic rendering effects and high scalability. 
(2) With the proposed techniques, the synthesis engine improves the performance of detectors and recognizers significantly. 
(3) We also generate a large scale multilingual scene text dataset that will aid further research. 
(4) Additionally, we notice that many of the popular scene text recognition datasets are only annotated in an incomplete way, providing only case-insensitive word annotations. 
With such limited annotations, researchers are unable to carry out comprehensive evaluations, and tend to overestimate the progress of scene text recognition algorithms. 
To address this issue, we re-annotate these datasets to include both \textit{upper-case} and \textit{lower-case characters}, \textit{digits}, \textit{punctuation marks}, and \textit{spaces} if there are any. 
We urge researchers to use the new annotations and evaluate in such a full-symbol mode for better understanding of the advantages and disadvantages of different algorithms.

\section{Related Work}
\subsection{Synthetic Images}
The synthesis of photo-realistic datasets has been a popular topic, since they provide detailed ground-truth annotations at multiple granularity, and cost less than manual annotations. 
In scene text detection and recognition, the use of synthetic datasets has become a standard practice. 
For scene text recognition, where images contain only one word, synthetic images are rendered through several steps~\cite{wang2012end,jaderberg2014synthetic}, including font rendering, coloring, homography transformation, and background blending. 
Later, GANs~\cite{goodfellow2014generative} are incorporated to maintain style consistency for implanted text~\cite{zhan2019spatial}, but it is only for single-word images. 
As a result of these progresses, synthetic data alone are enough to train state-of-the-art recognizers. 

To train scene text detectors, SynthText~\cite{gupta2016synthetic} proposes to generate synthetic data by printing text on background images. 
It first analyzes images with off-the-shelf models, and search suitable text regions on semantically consistent regions. 
Text are implanted with perspective transformation based on estimated depth. 
To maintain semantic coherency, VISD~\cite{zhan2018verisimilar} proposes to use semantic segmentation to filter out unreasonable surfaces such as human faces. 
They also adopt an adaptive coloring scheme to fit the text into the artistic style of backgrounds.
However, without considering the scene as a whole, these methods fail to render text instances in a photo-realistic way, and text instances are too outstanding from backgrounds. 
So far, the training of detectors still relies heavily on real images. 

Although GANs and other learning-based methods have also shown great potential in generating realistic images~\cite{wang2017adversarial,lin2018st,kar2019meta}, the generation of scene text images still require a large amount of manually labeled data~\cite{zhan2019spatial}. 
Furthermore, such data are sometimes not easy to collect, especially for cases such as low resource languages. 

More recently, synthesizing images with 3D graphics engine has become popular in several fields, including human pose estimation~\cite{varol2017learning}, scene understanding/segmentation~\cite{papon2015semantic,mccormac2016scenenet,richter2016playing,ros2016synthia,saleh2018effective}, and object detection~\cite{peng2015learning,tremblay2018falling,hinterstoisser2019annotation}. 
However, these methods either consider simplistic cases, e.g. rendering 3D objects on top of static background images~\cite{peng2015learning,varol2017learning} and randomly arranging scenes filled with objects~\cite{papon2015semantic,mccormac2016scenenet,ros2016synthia,hinterstoisser2019annotation}, or passively use off-the-shelf 3D scenes without further changing it~\cite{richter2016playing}. 
In contrast to these researches, our proposed synthesis engine implements active and regular interaction with 3D scenes, to generate realistic and diverse scene text images. 

This paper is also a sequel to our previous attempt, the SynthText3D\cite{liao2020synthtext3d}. 
SynthText3D closely follows the designs of the SynthText method. 
While SynthText uses off-the-shelf computer vision models to estimate segmentation and depth maps for background images, SynthText3D uses the ground-truth segmentation and depth maps provided by the 3D engines. 
The rendering process of SynthText3D does not involve interactions with the 3D worlds, such as the object meshes. 
As a result, SynthText3D is faced with at least these two limitations: (1) the camera locations and rotations are labeled by human, limiting the scalability as well as diversity; (2) the generated text regions are limited to well defined regions that the camera is facing upfront, resulting in a unfavorable location bias.

\subsection{Scene Text Detection and Recognition}
Scene text detection and recognition, possibly as the most human-centric computer vision task, has been a popular research topic for many years~\cite{ye2015text,long2018scene}. 
In {scene text detection}, there are mainly two branches of methodologies: {Top-down} methods that inherit the idea of region proposal networks from general object detectors that detect text instances as rotated rectangles and polygons~\cite{Liu2017Deep,Zhou_2017_CVPR,jiang2017r2cnn,zhang2019look,wang2019arbitrary}; {Bottom-up} approaches that predict local segments and local geometric attributes, and compose them into individual text instances~\cite{Shi_2017_CVPR,long2018textsnake,baek2019character,tian2019learning}. 
Despite significant improvements on individual datasets, those most widely used benchmark datasets are usually very small, with only around $500$ to $1000$ images in test sets, and are therefore prone to over-fitting. 
The generalization ability across different domains remains an open question, and is not studied yet. 
The reason lies in the very limited real data and that synthetic data are not effective enough. 
Therefore, one important motivation of our synthesis engine is to serve as a stepping stone towards general scene text detection. 

Most {scene text recognition} models consist of CNN-based image feature extractors and attentional LSTM~\cite{hochreiter1997long} or transformer~\cite{vaswani2017attention}-based encoder-decoder to predict the textual content~\cite{cheng2017arbitrarily,shi2018aster,li2018show,lyu20192d}. 
Since the encoder-decoder module is a language model in essence, scene text recognizers have a high demand for training data with a large vocabulary, which is extremely difficult for real-world data. 
Besides, scene text recognizers work on image crops that have simple backgrounds, which are easy to synthesize. 
Therefore, synthetic data are necessary for scene text recognizers, and synthetic data alone are usually enough to achieve state-of-the-art performance. Moreover, since the recognition modules require a large amount of data, synthetic data are also necessary in training \textbf{end-to-end text spotting} systems~\cite{liu2018fots,he2018end,qin2019towards}.

\section{Scene Text in 3D Virtual World} \label{sec:engine}
\subsection{Overview}
\begin{figure*}
\centering
      \includegraphics[width=0.9\linewidth]{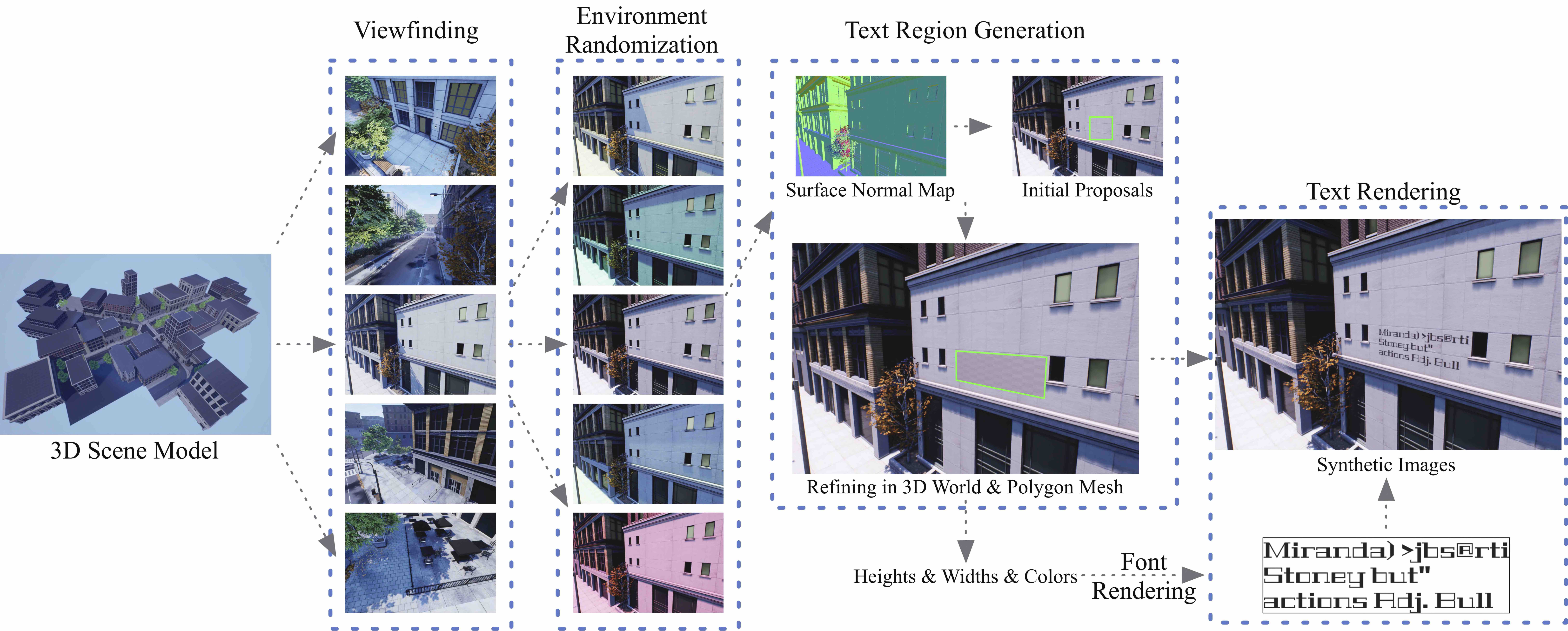}
    \caption{
    The pipeline of the proposed synthesis method. 
    The arrows indicate the order. 
    For simplicity, we only show one text region. 
    From left to right: scene overview, diverse viewpoints, various lighting conditions (light color, intensity, shadows, etc.), text region generation and text rendering.
    }
    \vspace{-5mm}
    \label{fig:pipeline}
\end{figure*}

In this section, we give a detailed introduction to our scene text image synthesis engine, \textit{UnrealText}, which is developed upon UE4 and the UnrealCV plugin~\cite{qiu2016unrealcv}. 
The synthesis engine: (1) produces \textbf{photo-realistic} images, (2) is \textbf{efficient}, taking about only $1$-$1.5$ second to render and generate a new scene text image and, (3) is \textbf{general and compatible} to off-the-shelf 3D scene models. 
As shown in Fig. ~\ref{fig:pipeline}, the pipeline mainly consists of a \textit{Viewfinder} module (section \ref{sec:view}), an \textit{Environment Randomization} module (section \ref{sec:env}), a \textit{Text Region Generation} module (section \ref{sec:reg}), and a \textit{Text Rendering} module (section \ref{sec:ren}). 

Firstly, the viewfinder module explores around the 3D scene with the camera, generating camera viewpoints. 
Then, the environment lighting is randomly adjusted. 
Next, the text regions are proposed based on 2D scene information and refined with 3D mesh information in the graphics engine. 
After that, text foregrounds are generated with randomly sampled fonts, colors, and text content, and are loaded as planar meshes. 
Finally, we retrieve the RGB image and corresponding text locations as well as text content to make the synthetic dataset. 

\subsection{Viewfinder} \label{sec:view}
\begin{figure}
\centering
      \includegraphics[width=1.0\linewidth]{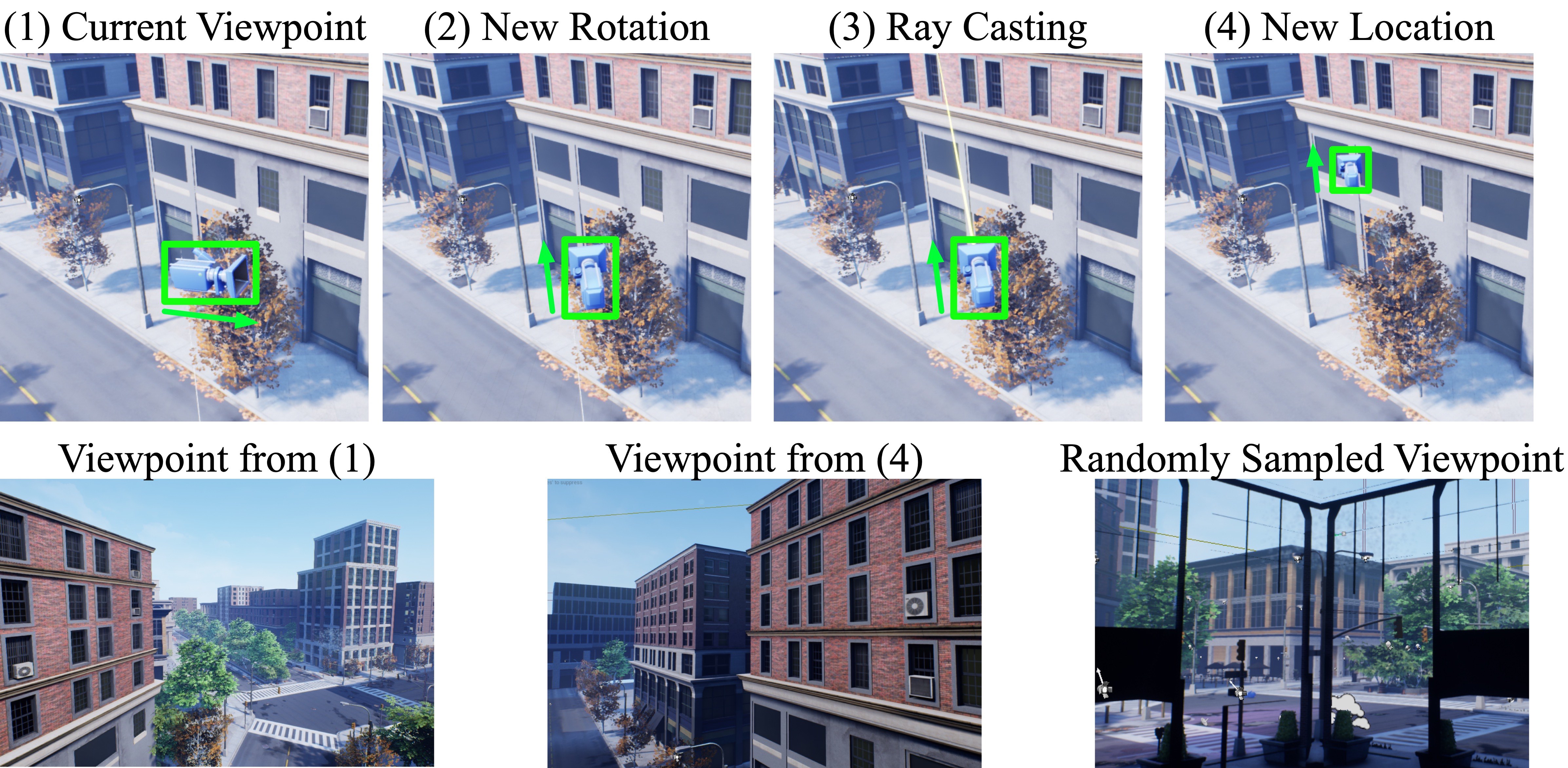}
      \vspace{-5mm}
    \caption{
    In the first row (1)-(4), we illustrate the \textit{physically-constrained 3D random walk}. 
    For better visualization, we use a camera object to represent the viewpoint (marked with \textcolor{green}{green boxes and arrows}). 
    In the second row, we compare viewpoints from the proposed method with randomly sampled viewpoints.
    }
    \label{fig:viewfinder}
\end{figure}

The aim of the viewfinder module is to automatically determine a set of camera locations and rotations from the whole space of 3D scenes that are reasonable and non-trivial, getting rid of unsuitable viewpoints such as from inside object meshes (e.g. Fig. ~\ref{fig:viewfinder} bottom right).  

Learning-based methods such as navigation and exploration algorithms may require extra training data and are not guaranteed to generalize to different 3D scenes. 
Therefore, we turn to rule-based methods and design a \textit{physically-constrained 3D random walk} (Fig. ~\ref{fig:viewfinder} first row) equipped with \textit{auxiliary camera anchors}. 

\subsubsection{Physically-Constrained 3D Random Walk}
Starting from a valid location, the physically-constrained 3D random walk aims to find the next valid and non-trivial location. 
In contrast to being valid, locations are invalid if they are inside object meshes or far away from the scene boundary, for example. 
A non-trivial location should be not too close to the current location. 
Otherwise, the new viewpoint will be similar to the current one. 
The proposed 3D random walk uses ray-casting~\cite{Roth1982Ray}, which is constrained by physically, to inspect the physical environment to determine valid and non-trivial locations. 

In each step, we first randomly change the pitch and yaw values of the camera rotation, making the camera pointing to a new direction. 
Then, we cast a ray from the camera location towards the direction of the viewpoint. 
The ray stops when it hits any object meshes or reaches a fixed maximum length. 
By design, the path from the current location to the stopping position is free of any barrier, i.e. not inside of any object meshes. 
Therefore, points along this ray path are all valid. 
Finally, we randomly sample one point between the $\frac{1}{3}$-th and $\frac{2}{3}$-th of this path, and set it as the new location of the camera, which is non-trivial. 
The proposed random walk algorithm can generate diverse camera viewpoints.

\subsubsection{Auxiliary Camera Anchors}
The proposed random walk algorithm, however, is inefficient in terms of exploration. 
Therefore, we manually select a set of $N$ camera anchors across the 3D scenes as starting points. 
After every $T$ steps, we reset the location of the camera to a randomly sampled camera anchor. 
We set $N=150$-$200$ and $T=100$.
Note that the selection of camera anchors requires only little carefulness. 
We only need to ensure coverage over the space. 
It takes around $20$ to $30$ seconds for each scene, which is trivial and not a bottleneck of scalability. 
The manual but efficient selection of camera is compatible with the proposed random walk algorithm that generates diverse viewpoints. 

\subsection{Environment Randomization} \label{sec:env}

To produce real-world variations such as lighting conditions, we randomly change the intensity, color, and direction of all light sources in the scene. 
In addition to illuminations, we also add fog conditions and randomly adjust its intensity. 
The environment randomization proves to increase the diversity of the generated images and results in stronger detector performance. 
The proposed randomization can also benefit sim-to-real domain adaptation~\cite{tobin2017domain}.

\subsection{Text Region Generation}  \label{sec:reg}
In real-world, text instances are usually embedded on well-defined surfaces, e.g. traffic signs, to maintain good legibility. 
Previous works find suitable regions by using estimated scene information, such as gPb-UCM~\cite{arbelaez2011contour} in SynthText~\cite{gupta2016synthetic} or saliency map in VISD~\cite{zhan2018verisimilar} for approximation. 
However, these methods are imprecise and often fail to find appropriate regions. 
Therefore, we propose to find text regions by probing around object meshes in 3D world. 
Since inspecting all object meshes is time-consuming, we propose a 2-staged pipeline: (1) We retrieve ground truth surface normal map to generate initial text region proposals; (2) Initial proposals are then projected to and refined in the 3D world using object meshes. 
Finally, we sample a subset from the refined proposals to render. To avoid occlusion among proposals, we project them back to screen space, and discard regions that overlap with each other one by one in a shuffled order until occlusion is eliminated. 

\subsubsection{Initial Proposals from Normal Maps}
In computer graphics, normal values are unit vectors that are perpendicular to a surface. 
Therefore, when projected to 2D screen space, a region with similar normal values tends to be a well-defined region to embed text on. 
We find valid image regions by applying sliding windows of $64\times64$ pixels across the surface normal map, and retrieve those with \textit{smooth} surface normal: the minimum cosine similarity value between any two pixels is larger than a threshold $t$. 
We set $t$ to $0.95$, which proves to produce reasonable results. 
We randomly sample at most $10$ non-overlapping valid image regions to make the initial proposals. 
Making proposals from normal maps is an efficient way to find potential and visible regions. 

\subsubsection{Refining Proposals in 3D Worlds}
\begin{figure}
\centering
      \includegraphics[width=1.0\linewidth]{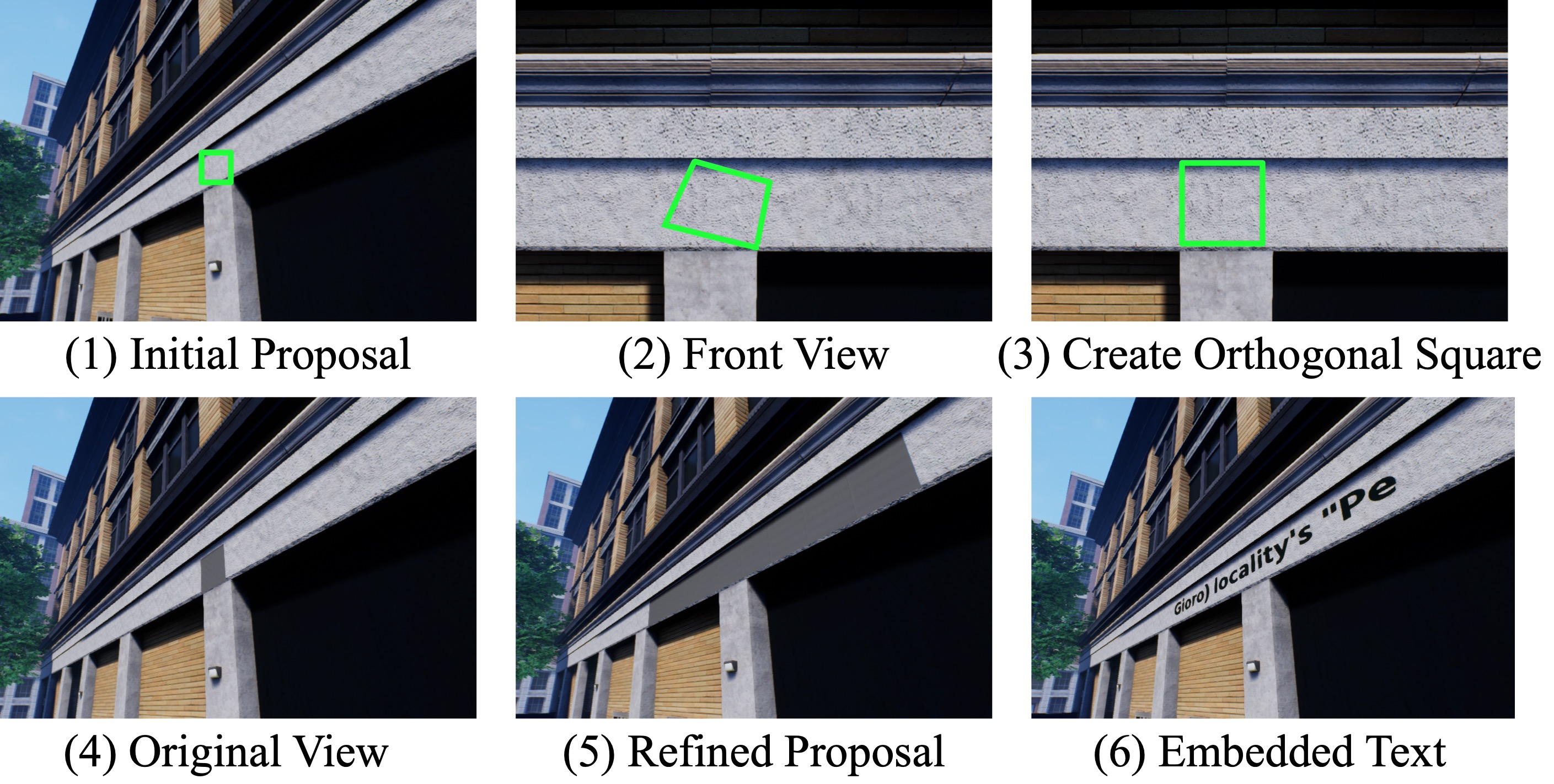}
      \vspace{-5mm}
    \caption{
    Illustration of the refinement of initial proposals. 
    We draw \textcolor{green}{green bounding boxes} to represent proposals in 2D screen space, and use planar meshes to represent proposals in 3D space. 
    (1) Initial proposals are made in 2D space. 
    (2) When we project them into 3D world and inspect them from the front view, they are in distorted forms. 
    (3) Based on the sizes of the distorted proposals and the positions of the center points, we re-initialize orthogonal squares on the same surfaces with horizontal sides orthogonal to the gravity direction. 
    (5) Then we expand the squares.  
    (6) Finally, we obtain text regions in 2D screen space with natural perspective distortion. 
    }
    \vspace{-3mm}
    \label{fig:refine}
\end{figure}

As shown in Fig. \ref{fig:refine}, 
rectangular initial proposals in 2D screen space will be distorted when projected into 3D world. 
Thus, we need to first rectify the proposals in 3D world. 
We project the center point of the initial proposals into 3D space, and re-initialize \textit{orthogonal} squares on the corresponding mesh surfaces around the center points: the horizontal sides are \textit{orthogonal} to the gravity direction. 
The side lengths are set to the shortest sides of the quadrilaterals created by projecting the four corners of initial proposals into the 3D space. 
Then we enlarge the widths and heights along the horizontal and vertical sides alternatively. 
The expansion of one direction stops when the sides of that direction get off the surface\footnote{when the distances from the rectangular proposals' corners to the nearest point on the underlying surface mesh exceed certain threshold}, hit other meshes, or reach the preset maximum expansion ratio. 
The proposed refining algorithm works in 3D world space, and is able to produce natural homography transformation in 2D screen space. 

\subsection{Text Rendering} \label{sec:ren}

\noindent \textbf{Generating Text Images:} Given text regions as proposed and refined in section ~\ref{sec:reg}, the text generation module samples text content and renders text images with certain fonts and text colors. 
The numbers of lines and characters per line are determined by the font size and the size of refined proposals in 2D space to make sure the characters are not too small and ensure legibility. 
For a fairer comparison, we also use the same font set from Google Fonts~\footnote{\url{https://fonts.google.com/}} as SynthText does. 
We also use the same text corpus, Newsgroup20. 
The generated text images have zero alpha values on non-stroke pixels, and non zero for others. 

\noindent \textbf{Rendering Text in 3D World:} We first perform triangulation for the refined proposals to generate planar triangular meshes that are closely attached to the underlying surface. 
Then we load the text images as texture onto the generated meshes. 
We also randomly sample the texture attributes, such as the ratio of diffuse and specular reflection. 

\subsection{Implementation Details}
The proposed synthesis engine is implemented based on UE4.22 and the UnrealCV plugin. 
On an ubuntu workstation with an 8-core Intel CPU, an NVIDIA GeForce RTX 2070 GPU, and 16G RAM, the synthesis speed is $0.7$-$1.5$ seconds per image with a resolution of $1080\times720$, depending on the complexity of the scene model. 

We collect $30$ scene models from the official UE4 marketplace. 
The engine is used to generate $600K$ 
scene text images with English words.   
With the same configuration, we also generate a multilingual version, making it the largest multilingual scene text dataset. 

\section{Experiments on Scene Text Detection
}

\subsection{Settings}
We first verify the effectiveness of the proposed engine by training detectors on the synthesized images and evaluating them on real image datasets. 
We use a previous yet time-tested state-of-the-art model, EAST~\cite{Zhou_2017_CVPR}, which is fast and accurate. 
EAST also forms the basis of several widely recognized end-to-end text spotting models~\cite{liu2018fots,he2018end}. 
We adopt an opensource implementation\footnote{\url{https://github.com/argman/EAST}}. 
In all experiments, models are trained on $4$ GPU with a batch size of $56$. 
During the evaluation, the test images are resized to match a short side length of $800$ pixels. 
For each experiment setting, we report the mean performance in $5$ independent trials. 

\noindent \textbf{Benchmark Datasets} We use the following scene text detection datasets for evaluation: (1) \textit{ICDAR 2013 Focused Scene Text} (IC13)~\cite{karatzas2013icdar} containing horizontal text with zoomed-in views. 
(2) \textit{ICDAR 2015 Incidental Scene Text} (IC15)~\cite{karatzas2015icdar} consisting of 
images taken without carefulness with Google Glass. Images are blurred and text are small. (3) \textit{MLT 2017}~\cite{nayef2017icdar2017} 
for multilingual scene text detection, which is composed of scene text images of $9$ languages. 
Note that the images in IC13 and MLT17 have varying resolutions. 
Therefore, it is necessary to resize them to the same level of resolutions before evaluation. 


\subsection{Experiments Results}
\noindent \textbf{Pure Synthetic Data} We first train the EAST models on different synthetic datasets alone, to compare our method with previous ones in a direct and quantitative way. 
Note that UnrealText, SynthText3D, SynthText, and VISD have different numbers of images, so we also need to control the number of images used in experiments. 
Results are summarized in Tab. \ref{tab:PSD}. 

Firstly, we control the total number of images to 10K, which is also the full size of the smallest synthetic datasets, VISD and SynthText3D. 
We observe a considerable improvement on IC15 over previous state-of-the-art by $+0.9\%$ in F1-score, and significant improvements on IC13 ($+2.7\%$) and MLT 2017 ($+2.8\%$). 
Secondly, we also train models on the full set of SynthText and ours, since scalability is also an important factor for synthetic scene text images, especially when considering the demand to train recognizers. 
Extra training images further improve F1 scores on IC15, IC13, and MLT by $+2.6\%$, $+2.3\%$, and $+2.1\%$.  
Models trained with our UnrealText data outperform all other synthetic datasets. 
Besides, the subset of $10K$ images with our method even surpasses $800K$ SynthText images significantly on all datasets. 
The experiment results demonstrate the effectiveness of our proposed synthetic engine and datasets. 

\begin{table}[ht]
\begin{center}
\begin{tabular}{|c|c|c|c|}
\hline

Training Data & IC15 & IC13 & MLT 2017   \\ \hline
SynthText 10K & 46.3 & 60.8 & 38.9   \\ 
VISD 10K (full) & 64.3 & 74.8 & 51.4   \\ 
SynthText3D 10K (full) & {{63.4}} & {{75.6}} & {48.3}   \\
UnrealText 10K & {\textbf{65.2}} & {\textbf{78.3}} & {\textbf{54.2}}   \\
\hline
SynthText 800K (full) & 58.0 & 67.7 & 44.8   \\ 
UnrealText 600K (full) & \textbf{67.8} & \textbf{80.6} & \textbf{56.3}   \\ \hline
\textit{SynthText3D 5K + VISD 5K} & \textit{65.4} & \textit{78.6} & \textit{52.2}\\
\textit{UnrealText 5K + VISD 5K} & \textit{\textbf{66.9}} & \textit{\textbf{80.4}} & \textit{\textbf{55.7}}\\ \hline
							
\end{tabular}
\end{center}
\vspace{-5mm}
\caption{Detection results (F1-scores) of EAST models trained on different synthetic data. 
}
\label{tab:PSD}
\end{table}

\noindent \textbf{Complementary Synthetic Data} One unique characteristic of the proposed UnrealText is that, the images are generated from 3D scene models, instead of real background images, resulting in potential domain gap due to different artistic styles. 
We conduct experiments by training on both UnrealText data ($5K$) and VISD ($5K$), as also shown in  Tab. \ref{tab:PSD} (last row, marked with \textit{italics}), which achieves better performance than other $10K$ synthetic datasets. 
The combination of UnrealText and VISD is also superior to the combination of SynthText3D and VISD. 
This result demonstrates that, our UnrealText is complementary to existing synthetic datasets that use real images as backgrounds. 
While UnrealText simulates photo-realistic effects, synthetic data with real background images can help adapt to real-world datasets.

\noindent \textbf{Combining Synthetic and Real Data} 
One important role of synthetic data is to serve as data for pretraining, and to further improve the performance on domain specific real datasets. 
We first pretrain the EAST models with different synthetic data, and then use domain data to finetune the models. 
The results are summarized in Tab. \ref{tab:finetune}. 
On all domain-specific datasets, models pretrained with our synthetic dataset surpasses others by considerable margins, verifying the effectiveness of our synthesis method in the context of boosting performance on domain specific datasets. 

\begin{table}[ht]
\begin{center}
\begin{tabular}{|c|c|c|c|}
\multicolumn{4}{c}{Evaluation on ICDAR 2015}   \\ \hline
Training Data & P & R & F1   \\ \hline
IC15 & 84.6 & 78.5 & 81.4	   \\ 
IC15 + SynthText 10K & 85.6 & 79.5 & 82.4	   \\ 
IC15 + VISD 10K & 86.3 & 80.0 & 83.1	   \\ 
IC15 + SynthText3D 10K & 86.6 & 80.4 & 83.4	   \\ 
IC15 + UnrealText 10K & \textbf{86.9} & \textbf{81.0} & \textbf{83.8}	   \\ \hline
\textit{IC15 + UnrealText 600K } & \textit{88.5} & \textit{80.8} & \textit{84.5}	   \\ 
\hline 
\multicolumn{4}{c}{Evaluation on ICDAR 2013}   \\ \hline
Training Data & P & R & F1   \\ \hline
IC13 & 82.6 & 70.0  & 75.8	   \\ 
IC13 + SynthText 10K & 85.3 & 72.4 & 78.3	   \\ 
IC13 + VISD 10K & 85.9 & 73.1 & 79.0	   \\ 
IC13 + SynthText3D 10K & 86.4 & 73.0 & 79.1   \\ 
IC13 + UnrealText 10K & \textbf{88.5} & \textbf{74.7} & \textbf{81.0}	   \\  \hline
\textit{IC13 + UnrealText 600K } & \textit{92.3} & \textit{73.4} & \textit{81.8}	   \\ 
\hline 
\multicolumn{4}{c}{Evaluation on MLT 2017}   \\ \hline
Training Data & P & R & F1   \\ \hline
MLT 2017 & 72.9 & 67.4 & 70.1	   \\ 
MLT 2017 + SynthText 10K & 73.1 & 67.7 & 70.3	   \\ 
MLT 2017 + VISD 10K & 73.3 & 67.9 & 70.5	   \\ 
MLT 2017 + SynthText3D 10K & 73.8 & 67.6 & 70.6	   \\ 
MLT 2017 + UnrealText 10K & \textbf{74.6} & \textbf{68.7} & \textbf{71.6}	   \\ \hline
\textit{MLT 2017 + UnrealText 600K } & \textit{{82.2}} & \textit{{67.4}} & \textit{{74.1}}	   \\ 
\hline

\end{tabular}
\end{center}
\vspace{-5mm}
\caption{Detection performances of EAST models pretrained on synthetic and then finetuned on real datasets. }
\label{tab:finetune}
\end{table}

\noindent \textbf{Pretraining on Full Dataset} As shown in the last rows of Tab. \ref{tab:finetune}, when we pretrain the detector models with our full dataset, the performances are improved significantly, demonstrating the advantage of the scalability of our engine.  
Especially, The EAST model achieves an F1 score of $74.1$ on MLT17, which is even better than recent state-of-the-art results, including $73.9$ by CRAFT\cite{baek2019character} and $73.1$ by LOMO \cite{zhang2019look}. 
Although the margin is not great, it suffices to claim that the EAST model revives and reclaims state-of-the-art performance with the help of our synthetic dataset. 

\noindent \textbf{Results with Mask-RCNN} As the EAST algorithm we use above is specifically designed for scene text and that the evaluation with F1 scores may not be comprehensive, we provide results with Mask-RCNN~\cite{he2017mask} which is a general object detector. 
We evaluate the models using the Average Precision (AP) metrics which are more comprehensive and less affected by the tricky choice of threshold values. 
We use the opensource implementation Detectron2~\footnote{\url{https://github.com/facebookresearch/detectron2}}. 
The rotated bounding boxes of text instances are used as the mask annotations. 
We select a default Mask-RCNN configuration with ResNet-50+FPN as the backbone and train the model for 1x schedule long. 
All the hyperparameters are set to default values. 
The results are summarized in Tab. \ref{tab:mask}. 
We notice that the two synthetic datasets with natural images as backgrounds, i.e. SynthText and VISD, result in similar performances. 
SynthText3D and our UnrealText are significantly better than them. 
UnrealText is further a significant improvement over SynthText3D. 
When we combine UnrealText and SynthText, the two highly scalable engines, the performances are even better.

\begin{table}[ht]
\begin{center}
\scalebox{0.78}{
\begin{tabular}{|c|c|c|c|}
\hline

Training Data & IC15 & IC13 & MLT 2017   \\ \hline
SynthText 10K & 13.6/13.4 & 41.1/41.7 & 19.6/17.5   \\ 
VISD 10K (full) & 13.8/13.5 & 37.6/37.6  & 18.1/18.4 \\ 
SynthText3D 10K (full)& {{19.4/19.8}} & {{37.9}}/{39.3} & 22.8/22.5   \\
UnrealText 10K & {\textbf{25.1}}/\textbf{23.7} & {\textbf{50.1}}/\textbf{49.2} & {\textbf{24}}/\textbf{23.6}   \\
\hline
SynthText 800K (full) & 19.6/20.3 & 47.5/48.2 & 24.2/24.8   \\ 
UnrealText 600K (full) & \textbf{26.2}/\textbf{25} & \textbf{51.5}/\textbf{52.2} & \textbf{27.8}/\textbf{27.3}   \\ \hline
\textit{UnrealText full + SynthText full} & \textit{\textbf{27.7}}/\textit{\textbf{27.5}} & \textit{\textbf{62.4}}/\textit{\textbf{63.7}} & \textit{\textbf{32.4}}/\textit{\textbf{32.1}}\\ \hline
							
\end{tabular}
}
\end{center}
\vspace{-5mm}
\caption{Detection results (Box-AP/Mask-AP) of Mask-RCNN models trained on different synthetic data. 
}
\label{tab:mask}
\end{table}

\subsection{Module Level Ablation Analysis} 
One reasonable concern about synthesizing from 3D virtual scenes lies in the scene diversity. 
In this section, we address the importance of the proposed view finding module and the environment randomization module in increasing the diversity of synthetic images. 

\noindent \textbf{Ablating Viewfinder Module} We derive two baselines from the proposed viewfinder module: (1) \textit{Random Viewpoint + Manual Anchor} that randomly samples camera locations and rotations from the norm-ball spaces centered around auxiliary camera anchors. 
(2) \textit{Random Viewpoint Only} that randomly samples camera locations and rotations from the whole scene space, without checking their quality. 
For experiments, we fix the number of scenes to $10$ to control scene diversity and generate different numbers of images, and compare their performance curve. 
By fixing the number of scenes, we compare how well different view finding methods can exploit the scenes.

 \noindent \textbf{Ablating Environment Randomization} 
 We remove the environment randomization module, and keep the scene models unchanged during synthesis. 
 For experiments, we fix the total number of images to $10K$ and use different number of scenes.
 In this way, we can compare the diversity of images generated with different methods. 
 
\begin{figure}
\centering
  \includegraphics[width=0.9\linewidth]{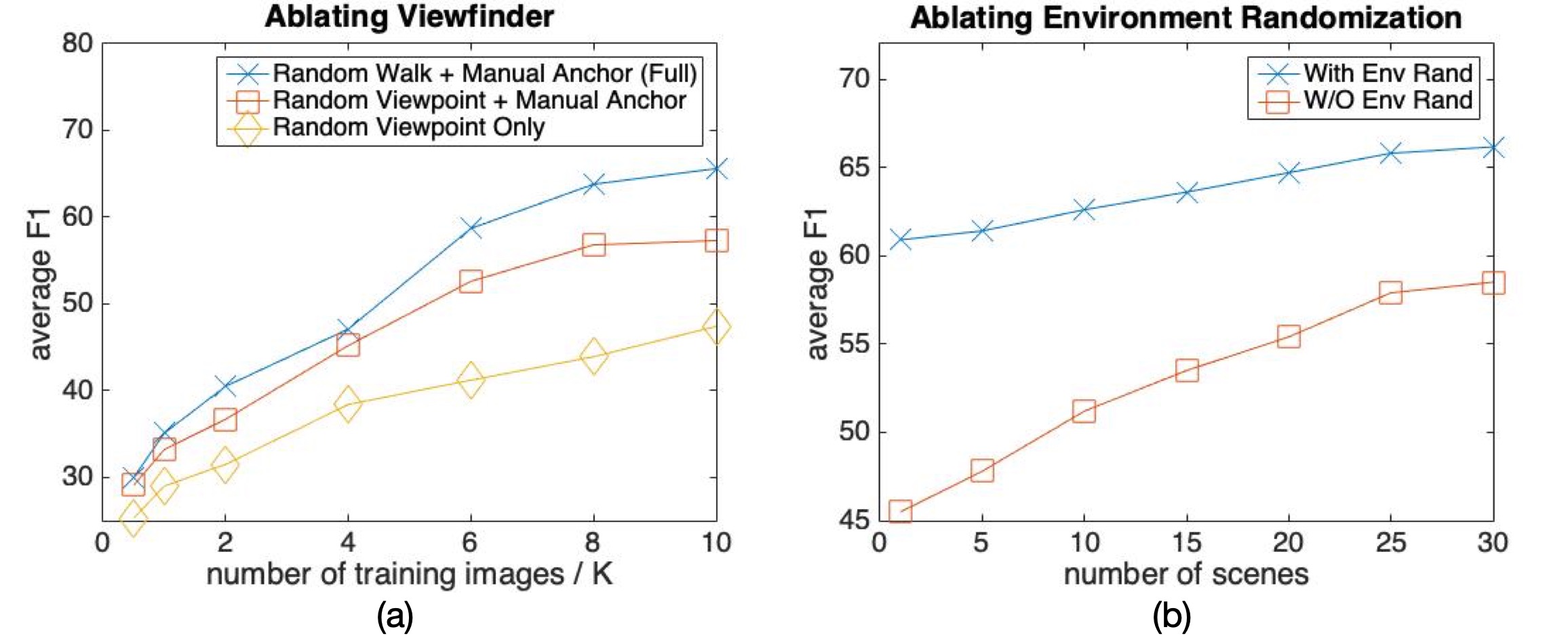}
  \vspace{-4mm}
\caption{Results of ablation tests: (a) ablating viewfinder module; (b) ablating environment randomization module. 
}
\label{fig:ablation}
\end{figure}

 We train the EAST models with different numbers of images or scenes, evaluate them on the $3$ real datasets, and compute the arithmetic mean of the F1-scores.   
 As shown in Fig. \ref{fig:ablation} (a), we observe that the proposed combination, i.e. \textit{Random Walk + Manual Anchor}, achieves significantly higher F1-scores consistently for different numbers of images. 
 Especially, larger sizes of training sets result in greater performance gaps. 
 We also inspect the images generated with these methods respectively. 
 When starting from the same anchor point, the proposed random walk can generate more diverse viewpoints 
 and can traverse much larger area. 
 In contrast, the \textit{Random Viewpoint + Manual Anchor} method degenerates either into random rotation only when we set a small norm ball size for random location, or into \textit{Random Viewpoint Only} when we set a large norm ball size. 
 As a result, the \textit{Random Viewpoint + Manual Anchor} method requires careful manual selection of anchors, and we also need to manually tune the norm ball sizes for different scenes, which restricts the scalability of the synthesis engine. 
Meanwhile, our proposed random walk based method is more flexible and robust to the selection of manual anchors. 
As for the \textit{Random Viewpoint Only} method, a large proportion of generated viewpoints are invalid, e.g. inside other object meshes, which is out-of-distribution for real images. 
This explains why it results in the worst performances.

 
 From Fig. \ref{fig:ablation} (b), the major observation is that environment randomization module improves performances over different scene numbers consistently. 
 Besides, the improvement is more significant as we use fewer scenes. 
 Therefore, we can draw a conclusion that, the environment randomization helps increase image diversity and at the same time, can reduce the number of scenes needed. 
 Furthermore, the random lighting conditions realize different real-world variations, which we also attribute as a key factor. 

\section{Experiments on Scene Text Recognition}
In addition to the superior performances in training scene text detection models, we also verify its effectiveness in the task of scene text recognition. 

\subsection{Recognizing Latin Scene Text}\label{rec-latin}

\subsubsection{Settings}
\noindent \textbf{Model} We select a widely accepted baseline method, ASTER~\cite{shi2018aster}, and adopt the implementation\footnote{\url{https://github.com/Jyouhou/ICDAR2019-ArT-Recognition-Alchemy}} that ranks top-1 on the ICDAR 2019 ArT competition on curved scene text recognition (Latin) by~\cite{long2019alchemy}. 
The models are trained with a batch size of $512$. 
A total of $95$ symbols are recognized, including an End-of-Sentence mark, $52$ case sensitive alphabets, $10$ digits, and $32$ printable punctuation symbols. 

\noindent \textbf{Training Datasets} From the $600K$ English synthetic images, we obtain a total number of $12M$ word-level image regions to make our training dataset. 
Also note that, our synthetic dataset provide character level annotations, which will be useful in some recognition algorithms. 

\noindent \textbf{Evaluation Datasets} 
We evaluate models trained on different synthetic datasets on several widely used real image datasets: \textit{IIIT}~\cite{mishra2012scene}, \textit{SVT}~\cite{wang2011end}, 
\textit{ICDAR 2015 (IC15)}~\cite{karatzas2015icdar},  \textit{SVTP}~\cite{quy2013recognizing},  \textit{CUTE}~\cite{risnumawan2014robust}, and \textit{Total-Text}\cite{kheng2017total}. 

Some of these datasets, however, have \textit{incomplete} annotations, including IIIT, SVT, SVTP, CUTE. 
While the word images in these datasets contain punctuation symbols, digits, upper-case and lower-case characters, the aforementioned datasets, in their current forms, only provide case-insensitive annotations and ignore all punctuation symbols. 
In order for more comprehensive evaluation of scene text recognition, we re-annotate these $4$ datasets in a case-sensitive way and also include punctuation symbols. 
We also release the new annotations and we believe that they will become better benchmarks for scene text recognition in the future.

\begin{table*}[ht]
\begin{center}
\scalebox{0.78}{
\begin{tabular}{|c|c|c|c|c|c|c|c|c|c|c|}
\hline
Training Data & Latin & Arabic  & Bangla & Chinese & Hindi & Japanese & Korean & Symbols & Mixed & Overall \\ \hline
ST (1.2M) & 34.6 & \textbf{50.5}  & \textbf{17.7} & 43.9 & 15.7 & 21.2 & \textbf{55.7} & \textbf{44.7} & 9.8 & 34.9 \\ 
UnrealText (1.2M) & \textbf{42.2} & 50.3 & 16.5 & \textbf{44.8} & \textbf{30.3} & \textbf{21.7} & 54.6 & 16.7 & \textbf{25.0} & \textbf{36.5} \\ 
\hline
\textit{UnrealText (full, 4.1M)} & {\textit{44.3}} & {\textit{51.1}} & {\textit{19.7}} & {\textit{47.9}} & {\textit{33.1}} & {\textit{24.2}} & {\textit{57.3}} & \textit{25.6} & {\textit{31.4}} & {\textit{39.5}} \\ 
\hline
MLT19-train (90K) & {64.3} & {47.2} & {46.9} & {{11.9}} & {46.9} & {23.3} & 39.1 & 35.9 & 3.6 & 45.7 \\ 
MLT19-train (90K) + ST (1.2M) & 63.8 & {62.0} & 48.9 & \textbf{50.7} & 47.7 & 33.9 & \textbf{64.5} & \textbf{45.5} & 10.3 & 54.7 \\ 
MLT19-train (90K) + UnrealText (1.2M) & \textbf{67.8} & \textbf{63.0} & \textbf{53.7} & 47.7 & \textbf{64.0} & \textbf{35.7} & 62.9 & 44.3 & \textbf{26.3} & \textbf{57.9} \\ 
\hline
\end{tabular}
}
\end{center}
\vspace{-5mm}
\caption{Multilingual scene text recognition results (word level accuracy).  
\textit{Latin} aggregates \textit{English}, \textit{French}, \textit{German}, and \textit{Italian}, as they are all marked as \textit{Latin} in the MLT dataset.
}
\label{tab:reg-mlt}
\end{table*}

\subsubsection{Experiment Results}
Experiment results are summarized in Tab. \ref{tab:reg}. 
First, we compare our method with previous synthetic datasets. 
We have to limit the size of training datasets to $1M$ since VISD only publishes $1M$ word images. 
Our synthetic data achieves consistent improvements on all datasets. 
Especially, it surpasses other synthetic datasets by a considerable margin on datasets with diverse text styles and complex backgrounds such as SVTP ($+2.4\%$). 
The experiments verify the effectiveness of our synthesis method in scene text recognition especially in the complex cases. 

Since small scale experiments are not very helpful in how researchers should utilize these datasets, we further train models on combinations of Synth90K, SynthText, and ours. 
We first limit the total number of training images to $9M$. 
When we train on a combination of all $3$ synthetic datasets, with $3M$ each, the model performs better than the model trained on $4.5M\times2$ datasets only. 
We further observe that training on $3M\times3$ synthetic datasets is comparable to training on the whole Synth90K and SynthText, while using much fewer training data. 
This result suggests that the best practice is to combine the proposed synthetic dataset with previous ones.

\begin{table}[ht]
\begin{center}
\scalebox{0.72}{
\begin{tabular}{|c|c|c|c|c|c|c|}
\hline
Training Data & IIIT & SVT  & IC15 & SVTP & CUTE & Total  \\ \hline
90K~\cite{jaderberg2014synthetic} (1M) & 51.6 & 39.2  & 35.7 & {37.2} & 30.9 & 30.5 \\ 
ST~\cite{gupta2016synthetic} (1M) & 53.5 & 30.3 & 38.4 & 29.5 & 31.2 & 31.1 \\ 
VISD~\cite{zhan2018verisimilar} (1M) & 53.9 & 37.1  & 37.1 & 36.3 & 30.5 & 30.9 \\ 
UnrealText (1M) & \textbf{54.8} & \textbf{40.3} & \textbf{39.1} & {\textbf{39.6}} & \textbf{31.6} & \textbf{32.1} \\ 
\hline
ST+90K($4.5M\times2$) & {80.5} & {70.1} & {58.4} & {{60.0}} & {63.9} & {43.2} \\ 
ST+90K+UnrealText($3M\times3$) & {\textbf{81.6}} & {\textbf{71.9}} & {{61.8}} & {{{61.7}}} & {\textbf{67.7}} & {\textbf{45.7}} \\ 
ST+90K($16M$) & {{81.2}} & {{71.2}} & {{\textbf{62.0}}} & {{{\textbf{62.3}}}} & {{65.1}} & {{44.7}} \\ 
\hline
\end{tabular}
}
\end{center}
\vspace{-5mm}
\caption{Results on English datasets (word level accuracy).  
}
\label{tab:reg}
\end{table}

\subsection{Recognizing Multilingual Scene Text}
\subsubsection{Settings}
Although MLT 2017 has been widely used as a benchmark for detection, the task of recognizing multilingual scene text still remains largely untouched, mainly due to lack of a proper training dataset. 
To pave the way for future research, we also generate a multilingual version with $600K$ images containing $10$ languages as included in MLT 2019~\cite{nayef2019icdar2019}: \textit{Arabic}, \textit{Bangla}, \textit{Chinese}, \textit{English}, \textit{French}, \textit{German}, \textit{Hindi}, \textit{Italian}, \textit{Japanese}, and \textit{Korean}. 
Text contents are sampled from corpus extracted from the Wikimedia dump\footnote{\url{https://dumps.wikimedia.org}}. 

\noindent \textbf{Model} We use the same model and implementation as Section \ref{rec-latin}, except that the symbols to recognize are expanded to all characters that appear in the generated dataset. 

\noindent \textbf{Training and Evaluation Data} We crop from the proposed multilingual dataset. 
We discard images with widths shorter than $32$ pixels as they are too blurry, and obtain $4.1M$ word images in total. 
We compare with the multilingual version of SynthText provided by \textit{MLT 2019} competition that contains a total number $1.2M$ images. 
For evaluation, we randomly split $1500$ images for each language (including \textit{symbols} and \textit{mixed}) from the training set of {MLT 2019}. 
The rest of the training set is used for training.

\subsubsection{Experiment Results}
Experiment results are shown in Tab. \ref{tab:reg-mlt}. 
When we only use synthetic data and control the number of images to $1.2M$, ours result in a considerable improvement of $1.6\%$ in overall accuracy, and significant improvements on some scripts, e.g. \textit{Latin} ($+7.6\%$) and \textit{Mixed} ($+21.6\%$). 
Using the whole training set of $4.1M$ images further improves overall accuracy to $39.5\%$.
When we train models on combinations of synthetic data and our training split of MLT19, as shown in the bottom of Tab. \ref{tab:reg-mlt}, we can still observe a considerable margin of our method over SynthText by $3.2\%$ in overall accuracy. 
The experiment results demonstrate that our method is also superior in multilingual scene text recognition, and we believe this result will become a stepping stone to further research.

\section{Limitation and Future Work}
There are several aspects that are worth diving deeper into: (1) Overall, the engine is based on rules and human-selected parameters. The automation of the selection and search for these parameters can save human efforts and help adapt to different scenarios. 
(2) While rendering small text can help training detectors, the low image quality of the small text makes recognizers harder to train and harms the performance. 
Designing a method to mark the illegible ones as \textit{difficult} and excluding them from loss calculation may help mitigate this problem. 
(3) For multilingual scene text, scripts except \textit{Latin} have much fewer available fonts that we have easy access to. 
To improve performance on more languages, researchers may consider learning-based methods to transfer \textit{Latin} fonts to other scripts. 

\section{Conclusion}
In this paper, we introduce a scene text image synthesis engine that renders images with 3D graphics engines, where text instances and scenes are rendered as a whole. 
In experiments, we verify the effectiveness of the proposed engine in both scene text detection and recognition models. 
We also study key components of the proposed engine. 
We believe our work will be a solid stepping stone towards better synthesis algorithms.

\section*{Acknowledgement}
\vspace{-2mm}
This research was supported by National Key R\&D Program of China (No. 2017YFA0700800).

{\small
\bibliographystyle{ieee_fullname}
\bibliography{egbib}
}

\newpage

\newpage

\section*{A. Scene Models}

In this work, we use a total number of $30$ scene models which are all obtained from the Internet. 
However, most of these models are not free. 
Therefore, we are not allowed to share the models themselves. 
Instead, we list the models we use and their links in Tab. \ref{tab:scenes}. 

\begin{table*}
\begin{center}
\scalebox{0.78}{
\begin{tabular}{|c|c|}
\hline
Scene Name & Link\\ \hline
Urban City & \url{https://www.unrealengine.com/marketplace/en-US/product/urban-city}\\ \hline
Medieval Village & \url{https://www.unrealengine.com/marketplace/en-US/product/medieval-village}\\ \hline
Loft & \url{https://ue4arch.com/shop/complete-projects/archviz/loft/}\\ \hline
Desert Town & \url{https://www.unrealengine.com/marketplace/en-US/product/desert-town}\\ \hline
Archinterior 1 & \url{https://www.unrealengine.com/marketplace/en-US/product/archinteriors-vol-2-scene-01} \\ \hline
Desert Gas Station & \url{https://www.unrealengine.com/marketplace/en-US/product/desert-gas-station} \\ \hline
Modular School & \url{https://www.unrealengine.com/marketplace/en-US/product/modular-school-pack}\\ \hline
Factory District & \url{https://www.unrealengine.com/marketplace/en-US/product/factory-district} \\ \hline
Abandoned Factory & \url{https://www.unrealengine.com/marketplace/en-US/product/modular-abandoned-factory}\\ \hline
Buddhist & \url{https://www.unrealengine.com/marketplace/en-US/product/buddhist-monastery-environment}\\ \hline
Castle Fortress & \url{https://www.unrealengine.com/marketplace/en-US/product/castle-fortress}\\ \hline
Desert Ruin & \url{https://www.unrealengine.com/marketplace/en-US/product/modular-desert-ruins}\\ \hline
HALArchviz & \url{https://www.unrealengine.com/marketplace/en-US/product/hal-archviz-toolkit-v1} \\ \hline
Hospital & \url{https://www.unrealengine.com/marketplace/en-US/product/modular-sci-fi-hospital}\\ \hline
HQ House & \url{https://www.unrealengine.com/marketplace/en-US/product/hq-residential-house}\\ \hline
Industrial City & \url{https://www.unrealengine.com/marketplace/en-US/product/industrial-city}\\ \hline
Archinterior 2 & \url{https://www.unrealengine.com/marketplace/en-US/product/archinteriors-vol-4-scene-02}\\ \hline
Office & \url{https://www.unrealengine.com/marketplace/en-US/product/retro-office-environment}\\ \hline
Meeting Room & \url{https://drive.google.com/file/d/0B\_mjKk7NOcnEUWZuRDVFQ09STE0/view} \\ \hline
Old Village & \url{https://www.unrealengine.com/marketplace/en-US/product/old-village}\\ \hline
Modular Building & \url{https://www.unrealengine.com/marketplace/en-US/product/modular-building-set}\\ \hline
Modular Home & \url{https://www.unrealengine.com/marketplace/en-US/product/supergenius-modular-home}\\ \hline
Dungeon & \url{https://www.unrealengine.com/marketplace/en-US/product/top-down-multistory-dungeons}\\ \hline
Old Town & \url{https://www.unrealengine.com/marketplace/en-US/product/old-town}\\ \hline
Root Cellar & \url{https://www.unrealengine.com/marketplace/en-US/product/root-cellar}\\ \hline
Victorian & \url{https://www.unrealengine.com/marketplace/en-US/product/victorian-street}\\ \hline
Spaceship & \url{https://www.unrealengine.com/marketplace/en-US/product/spaceship-interior-environment-set}\\ \hline
Top-Down City & \url{https://www.unrealengine.com/marketplace/en-US/product/top-down-city}\\ \hline
Scene Name & \url{https://www.unrealengine.com/marketplace/en-US/product/urban-city}\\ \hline
Utopian City & \url{https://www.unrealengine.com/marketplace/en-US/product/utopian-city}\\ \hline
\end{tabular}
}
\end{center}
\vspace{-5mm}
\caption{The list of 3D scene models used in this work. }
\label{tab:scenes}
\end{table*}

\section*{B. New Annotations for Scene Text Recognition Datasets}
During the experiments of scene text recognition for English scripts, we notice that among the most widely used benchmark datasets, several have incomplete annotations. 
They are IIIT5K, SVT, SVTP, and CUTE-80. 
The annotations of these datasets are case-insensitive, and ignore punctuation marks. 

The common practice for recent scene text recognition research is to convert both prediction and ground-truth text strings to lower-case and then compare them. 
This means that the current evaluation is flawed. 
It ignores letter case and punctuation marks which are crucial to the understanding of the text contents. 
Besides, evaluating on a much smaller vocabulary set results in over-optimism of the performance of the recognition models. 

To aid further research, we use the Amazon mechanical Turk (AMT) to re-annotate the aforementioned $4$ datasets, which amount to $6837$ word images in total. 
Each word image is annotated by $3$ workers, and we manually check and correct images where the $3$ annotations differ. 
The annotated datasets are released via GitHub at \url{https://github.com/Jyouhou/Case-Sensitive-Scene-Text-Recognition-Datasets}.

\subsection*{B.1 Samples}
\begin{figure}
\centering
      \includegraphics[width=1.0\columnwidth]{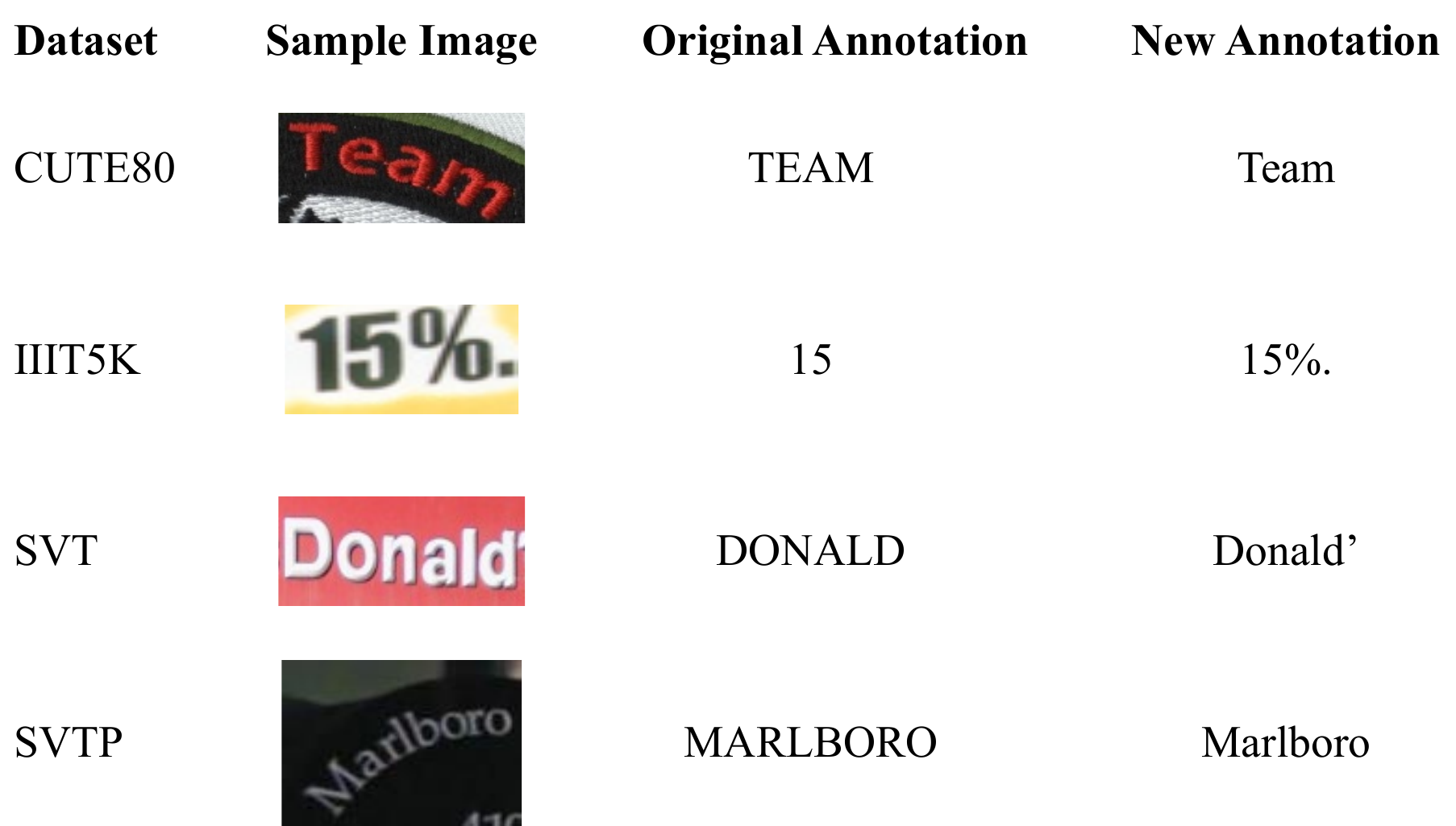}
    \caption{
    Examples of the new annotations. 
    }
    \label{fig:new}
\end{figure}

We select some samples from the $4$ datasets to demonstrate the new annotations in Fig. \ref{fig:new}.

\subsection*{B.2 Benchmark Performances}
As we are encouraging case-sensitive (also with punctuation marks) evaluation for scene text recognition, we would like to provide benchmark performances on those widely used datasets. 
We evaluate two implementations of the ASTER models, by Long \emph{et al.}\footnote{\url{https://github.com/Jyouhou/ICDAR2019-ArT-Recognition-Alchemy}} and Baek \emph{et al}\footnote{\url{https://github.com/clovaai/deep-text-recognition-benchmark}} respectively. 
Results are summarized in Tab. \ref{tab:reg}. 

The two benchmark implementations perform comparably, with Baek's better on straight text and Long's better at curved text. 
Compared with evaluation with \textit{lower case + digits}, the performance drops considerably for both models when we evaluate with all symbols. 
These results indicate that it may still be a challenge to recognize a larger vocabulary, and is worth further research. 

\begin{table*}[ht]
\begin{center}
\scalebox{1.0}{
\begin{tabular}{|c|c|c|c|c|c|c|c|c|}
\hline
Implementation & Case & IIIT & SVT & IC13 & IC15 & SVTP & CUTE80 & Total \\ \hline
Long \emph{et al.} & All & {81.2} & {71.2} & {86.9} & {{62.0}} & {62.3} & {65.1} & 44.7 \\ 
Baek \emph{et al} & All & {{81.5}} & {{71.7}} & {{88.9}} & {{{62.1}}} & {{62.6}} & {{64.9}} & 41.5\\ 
\hline
Long \emph{et al.} & lower case + digits & {89.5} & {84.1} & {89.9} & {{68.8}} & {73.5} & {76.3} & 58.2 \\ 
Baek \emph{et al} & lower case + digits & {{86.5}} & {{83.5}} & {{93.0}} & {{{70.3}}} & {{75.1}} & {{68.4}} & 46.0\\ 
\hline
\end{tabular}
}
\end{center}
\vspace{-5mm}
\caption{Results on English datasets (word level accuracy).  
\textit{All} indicates that the evaluation considers lower case characters, upper case characters, numerical digits, and punctuation marks. 
}
\label{tab:reg}
\end{table*}

\newpage
\newpage

\end{document}